\documentclass{article}

\usepackage{neurips_2023}

\usepackage[utf8]{inputenc} 
\usepackage[T1]{fontenc}    
\usepackage{hyperref}       
\usepackage{url}            
\usepackage{booktabs}       
\usepackage{amsfonts}       
\usepackage{nicefrac}       
\usepackage{microtype}      
\usepackage{xcolor}         
\usepackage{mwe}
\usepackage{float}  
\usepackage{graphicx} 
\usepackage{tikz}

\usepackage{url}
\usepackage{multirow}
\usepackage{adjustbox}
\usepackage{footnote}
\usepackage{booktabs}
\usepackage{graphicx}
\usepackage[none]{hyphenat}
\usepackage{makecell}
\usepackage{framed}
\usepackage{tabto}
\usepackage{hyperref}
\usepackage{tipa}
\usepackage{subcaption}

\usepackage{amsmath}

\usepackage{ulem}
\usepackage{makecell}

\def\SSigma{\boldsymbol{\Sigma}}

\title{SwiftLearn: A Data-Efficient Training Method of Deep Learning Models using Importance Sampling}

\author{\textbf{Habib Hajimolahoseini, Omar Mohamed Awad, Walid Ahmed,} \\ \textbf{Austin Wen, Saina Asani, Mohammad Hassanpour, Farnoosh Javadi,} \\ \textbf{Mehdi Ahmadi, Foozhan Ataiefard, Kangling Liu, Yang Liu}\\
$^*$Ascend Team, Toronto Research Center, Huawei Technologies \\
  \texttt {habib.hajimolahoseini@huawei.com}\\
}

\begin{document}

\maketitle

\begin{abstract}
    In this paper, we present \textit{SwiftLearn}, a data-efficient approach to accelerate training of deep learning models using a subset of data samples selected during the warm-up stages of training.
    This subset is selected based on an importance criteria measured over the entire dataset during warm-up stages, aiming to preserve the model performance with fewer examples during the rest of training. 
    The importance measure we propose could be updated during training every once in a while, to make sure that all of the data samples have a chance to return to the training loop if they show a higher importance. 
    The model architecture is unchanged but since the number of data samples controls the number of forward and backward passes during training, we can reduce the training time by reducing the number of training samples used in each epoch of training.
    Experimental results on a variety of CV and NLP models during both pretraining and finetuning show that the model performance could be preserved while achieving a significant speed-up during training. More specifically, BERT finetuning on GLUE benchmark shows that almost 90\% of the data can be dropped achieving an end-to-end average speedup of 3.36$\times$ while keeping the average accuracy drop less than 0.92\% .


\end{abstract}

\section{Introduction}
\label{sec:intro}

Over the past few years, deep learning models have undergone a significant increase in size, boasting millions, or even billions, of trainable parameters \citep{li2021short, radford2018improving, awad2023improving, hajimolahoseini2023training, hajimolahoseini2023methods}. 
The training of such immense models requires substantial memory resources and computational power \citep{brown2020language, openai2023gpt4, touvron2023llama, hajimolahoseini2018ecg, hajimolahoseini2022long, hajimolahoseini2012extended}. 
Furthermore, they demand extensive training on large datasets, resulting in prolonged training durations. 
These challenges impose restrictions on deploying these models on edge devices with limited computational capabilities, thereby limiting their applicability across numerous scenarios.

To tackle these issues, three distinct categories of methods have been introduced to accelerate the training either by Structural Efficiency, Data Efficiency, or Hardware Efficiency. 
Structural efficiency improvements involve modifications to the model's architecture to enhance computational efficiency, essentially reducing its computational or memory demands. 
Some notable examples of such methods encompass network pruning \citep{sun2023simple}, low-rank decomposition \citep{hu2021lora, hajimolahoseini2022strategies, hajimolahoseini2016robust, li2021short, ahmed2023speeding}, and weight quantization \citep{dettmers2023qlora, hajimolahoseini2018inflection, hajimolahoseini2019deep}, among others.
Hardware-efficient training, on the other hand, focuses on achieving high model performance while minimizing the hardware requirements \citep{dao2022flashattention}, which can be crucial for practical deployment in resource-constrained environments, such as edge devices, mobile phones, and embedded systems.

In contrast, data-efficient training approaches do not change the model's architecture. 
Instead, they focus on expediting the training process by excluding less crucial data samples from the training dataset \citep{okanovic2023repeated}. In fact, since the number of data samples directly influences the number of forward and backward passes during training, reducing the number of training samples used in each training epoch can substantially reduce the overall training time.
This paper primarily concentrates on this category of techniques.
The key advantage of this approach lies in its independence from the model's structure or the specific task at hand. Consequently, any successful data-efficient method can be employed concurrently with other techniques aimed at expediting the training process, providing a synergistic boost to overall training speed.

Data-efficient training methods can be classified into three main categories: Dataset Condensation, Dataset Pruning, and Curriculum Learning. Each of these categories aims to optimize the use of training data in deep learning models. Dataset Condensation involves the synthesis of a new, smaller dataset that captures the essential distributional characteristics of the original dataset. This new dataset may not necessarily contain the same data samples as the original one, but its distribution closely aligns with that of the complete dataset. Techniques used for dataset condensation include optimizing Kernel ridge-regression with respect to input data samples \citep{10.1145/3580305.3599398}, employing generator functions \citep{zhao2021dataset}, utilizing auxiliary small unlabeled samples \citep{sattler2021fedaux}, and learning soft labels on a reduced dataset \citep{vyas2020learning}.

Dataset Pruning on the other hand focuses on reducing the size of the original dataset by removing less important and less representative data samples. Unlike dataset condensation, the resulting dataset in this case is a subset of the original data. Recent approaches have introduced novel metrics, such as gradient norms, to determine the importance of each data sample \citep{ahn2023mitigating}. Alternatively, a scaled-down model can be trained to serve as a proxy for selecting important data samples based on predefined metrics \citep{coleman2020selection}. Other metrics include the number of times a model misclassifies a sample and the high variance of gradients, indicating example difficulty \citep{dadalto2023datadriven}. Prediction depth, the layer at which a k-NN classifier can successfully classify an example, can also serve as a measure of computational difficulty \citep{ma2018sparsetodense}.

Curriculum learning methods revolve around identifying the optimal order of data samples to accelerate model convergence. Unlike dataset pruning, these methods do not remove data samples but emphasize the importance of the order in which data samples are presented during training. Curriculum learning methods define metrics to quantify the importance of each data sample, and the order of learning is determined based on these scores. These metrics can be crafted manually or generated automatically \citep{zhang2021learning, demathelin2022fast}. While handcrafted metrics may be task-specific, they often outperform automated metrics. Nevertheless, both types of metrics remain valuable for various tasks. It is worth noting that most dataset reduction techniques have primarily been explored in image classification tasks, with limited exploration in text-based tasks.

In this paper, we propose \textit{SwiftLearn}, a technique that could be considered as a combination of dataset pruning and curiculum learning. The proposed algorithm is described in the following sections.

\section{Proposed Algorithm}
The proposed method is applied in two steps: Warm-up and Sampling:

\subsection{Warm-up}
In the warm-up step, which happens during the first 2 epochs of each training phase, all data samples in the dataset are fed to the model. 
During this step, the model logits are recorded in order to be used later for calculating the importance metric. 
The Mean Squared Error (MSE) between the logits of the model in two consecutive epochs is calculated for each data sample $x_i$ as follows:
\begin{equation}\label{mse}
\text{MSE($x_i$)} = ||P_i^s - P_i^{s-1}||_2
\end{equation}
where $P_i^s$ is the logits of the model for data sample $x_i$ at epoch $s$ and $P_i^{s-1}$ is the logits of the model for the same sample in the previous epoch $s-1$. This metric shows how the logits of the model are changing between consecutive epochs of training. This measure is then normalized using a softmax function in order to make it a probability distribution:
\begin{equation}\label{softmax}
\text{Pr($x_i$)} = \frac{e^{(\sigma\times\text{MSE($x_i$)})}}{\SSigma_j e^{(\sigma\times\text{MSE($x_j$)})}}
\end{equation}
where $\sigma$ is a controllable temperature factor. 
This probability is used in the next stage as a measure of importance, based on which the data samples are chosen as a subset for training. 
Note that the warm-up stage is applied during the training of the model (not before that) and no data sample is removed in this stage. 
We give all the samples a chance to contribute to the training but with different probability of being chosen.

\begin{figure*}[h!]
    \centering
    \includegraphics[width=0.8\textwidth]{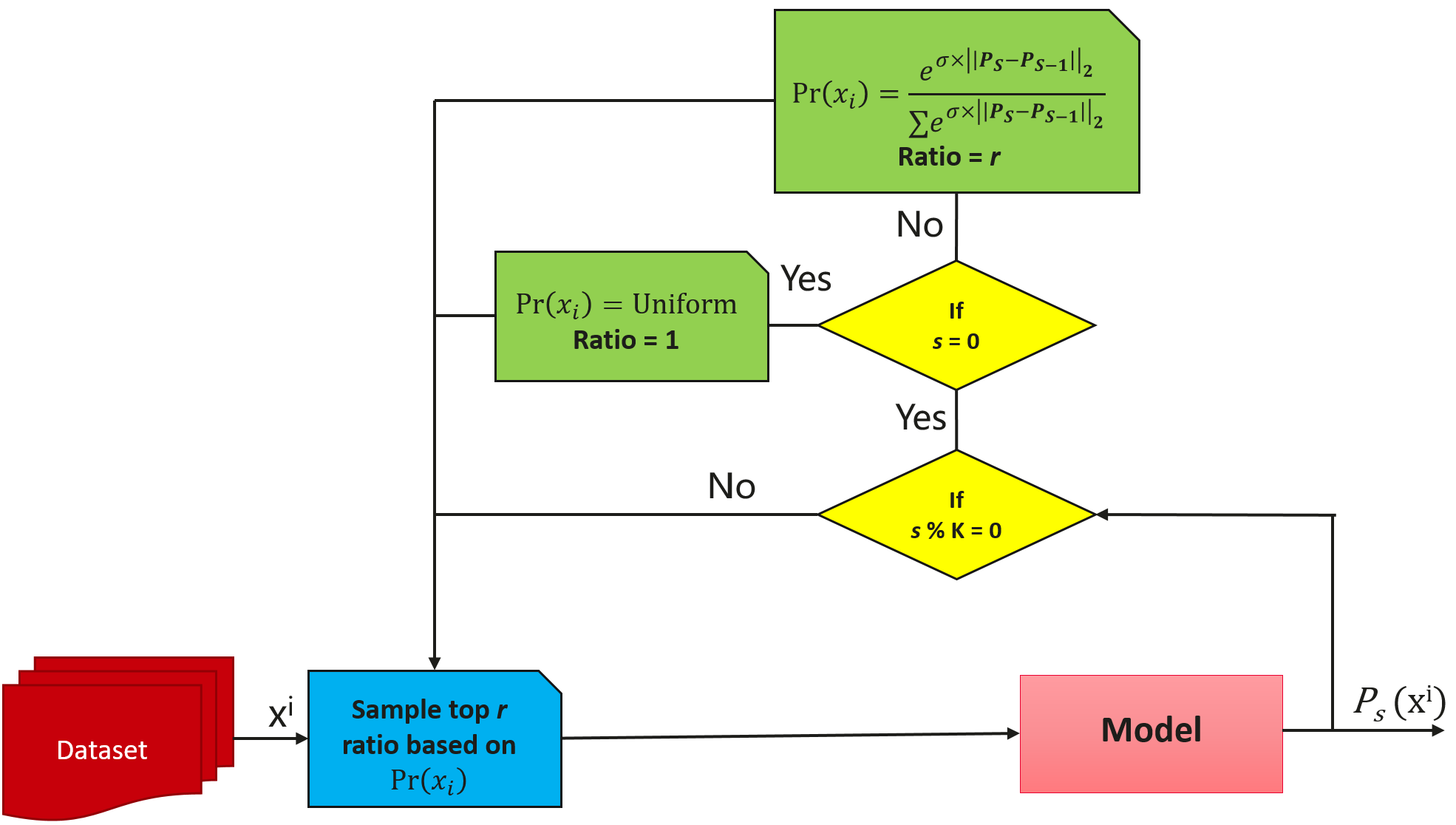}
    \caption{The proposed data efficient sampling technique}
    \label{fig:alg}
    \vspace*{-3mm}
\end{figure*}

\subsection{Sampling}
After the sampling metric is calculated/updated in the previous step, this measure is used to sample from the dataset using a ratio $r$ defined by the user. 
This speed-up is controlled by two factors: ratio $r$ and Metric Re-evaluation Interval. 
The ratio $r$ controls the number of samples fed through the network(s). 
The lower the ratio, the less number of data samples are chosen from dataset. 
If $N$ is the total number of data samples in the dataset, the number of data samples that will be fed during each epoch will be $N \times r$ where $0\leq r \leq 1$. By feeding only a portion of data-set, the number of forward and backward passes decreases which leads to faster training. 

On the other hand, the metric that we find during the warm-up stage may change during training, if the network’s status is affected by that. To get the maximum speed up, we need to stick to the metric that we find or we can update that during each epoch for only those samples which have been chosen. This partial or lack of update causes degradation in performance. We can sacrifice the speed-up to perform better by re-evaluating the metric frequently for the whole data-set. By doing that, the forward pass needs to be done for all samples frequently. The length of this re-evaluation interval controls the save in time that our algorithm provides.

The entire data efficient sampling process is depicted in Fig. \ref{fig:alg}. 
In this figure, $K$ represents the frequency by which we update the importance metric. 
At the $0^{th}$ epoch, all of the samples from the original model are used (ratio $r$ = 1). 
Every $K^{th}$ epoch, the importance function $Pr$ is calculated using the softmax equation and MSE function. In our experiments, we set $K$ to be the total number of training epochs so the importance metric is set only once for simplicity. Then, the top $r$ number of samples from the dataset are chosen according to the importance metric. This process is repeated until the model is fully trained.

\begin{table*}[h!]
\centering
\setlength{\tabcolsep}{3pt}
\resizebox{\linewidth}{!}{
\begin{tabular}{lcccc|cccc}
\toprule
\multirow{2}{*}{Model} & \multicolumn{1}{c}{\textbf{Task}} & \multicolumn{1}{c}{\textbf{Dataset}} & \multicolumn{1}{c}{\textbf{Drop Ratio}} & \multicolumn{1}{c}{\textbf{Speed-up}} & \multicolumn{4}{c}{\textbf{Performance}} \\
 & & & & & Data-weighting & Baseline & Metric & $\Delta$ Acc \\
\midrule
T5 & Finetuning & wmt16/en-ro & 0.3 & $1.08 \times$ & 18.44 & 18.46 & Bleu & -0.02 \\ \midrule
RoBERTa & Finetuning & SST-2 & 0.3 & 1.35$ \times$ & 94.00 & 94.00 & Acc $\%$ & 0.00\\ \midrule
ViT & Pretraining & ImageNet & 0.1 & 1.05$\times$ & 69.89 & 70.01 & Acc $\%$ & -0.12 \\ \midrule
SwinT & Pretraining & ImageNet  & 0.2 & 1.136$\times$ & 81.00 & 81.00 & Acc $\%$ & 0.00 \\ \midrule
Conformer & Pretraining & LibriSpeech & 0.3 & 1.338$\times$ & 95.07 & 95.32 & Acc $\%$ & -0.25 \\ \midrule
BERT-Large & Pretraining & Chinese-wiki & 0.2 & 1.245$\times$ & 70.68 & 71.43 & Acc $\%$ & -0.75 \\

\bottomrule

\end{tabular}
}
\caption{Comparison between the accuracy and speed-up of data-weighting and the vanilla training baseline across multiple tasks and models.} 
\vspace{-3mm}
\label{tab:results}
\end{table*}

\section{Experimental Results}
\textbf{Comparisons with baselines}. Here, we evaluate \textit{SwiftLearn} over pretraining and finetuning tasks on a variety of large deep learning models as shown in table~\ref{tab:results}. 
In our experiments, we trained the models on Huawei's Ascend 910 distributed over 8 devices. 
Across the models we studied, the SwinT \citep{liu2021swin}, Conformer \citep{gulati2020conformer}, and RoBERTa \citep{liu2019roberta} models showed the best speedup-accuracy tradeoff. We pretrained SwinT on the ImageNet dataset \citep{5206848} for 300 epochs with a drop ratio of 0.2 and we achieved an end-to-end (E2E) speedup of $13.6\%$ with no loss in the final validation accuracy. We also pretrained Conformer on the LibriSpeech \citep{panayotov2015librispeech} dataset for 50 epochs with a drop ratio of 0.3 achieving a $33.8\%$ E2E speedup with only a $-0.25\%$ drop in the validation accuracy. We finetuned RoBERTa on the SST-2 dataset for 10 epochs with a drop ratio of 0.3 achieving a $35\%$ E2E speedup with no loss in the final validation accuracy. This confirms our hypothesis that not all training samples contribute equally to the model's learning and that with a careful selection of important samples, only a smaller subset will be sufficient.

Although finetuning BERT-Large \citep{devlin2019bert} on the Chinese-wiki \citep{xu2019weakly} dataset achieved a $24.5\%$ E2E speedup with a drop ratio of 0.2, it suffered from a relatively higher drop in the validation accuracy of $-0.75\%$. This is because Chinese-wiki is a small Pretraining dataset with only 1.2M samples and is trained using only 3 epochs
Since the first two epochs are used during the warm-up stage, it may need more epochs during the sampling step to result in a higher accuracy. 

Finetuning the T5 \citep{2020t5} model on the WMT 2016 \citep{bojar-EtAl:2016:WMT1} dataset for 3 epochs using a drop ratio of 0.3 achieved an 8$\%$ E2E speedup with a slight decrease in the Bleu score of 0.02. 

\textbf{Effect of drop ratio on speed up and accuracy}. Table~\ref{tab:BERT_results} shows how sweeping on the drop ratio affects the E2E speedup and accuracy of finetuning BERT on different tasks from the GLUE benchmark \citep{wang2019glue}. The experiments are done on Huawei's Ascend 910 chips. All studied tasks showed significant improvement in the E2E speedup up to $3.52\times$. CoLA, SST-2, and QNLI tasks showed an increase in the validation accuracy with drop ratios of 0.9 and 0.7 relative to the baseline of 0.9-4$\%$, and 1.4-4.1$\%$, respectively. SST-2 with a drop ratio of 0.7 showed the highest accuracy improvement with $4.1\%$ increase in the validation accuracy relative to the baseline. Only for MRPC that we saw a significant drop in the validation accuracy of $10.2\%$ and $10.5\%$ for a drop ratio of 0.9 and 0.7, respectively. This can be attributed to MRPC being a relatively more difficult task that requires a much lower drop ratio to maintain the baseline accuracy.
However, in general, BERT finetuning on GLUE benchmark shows that almost 90\% of the data can be dropped while keeping the average accuracy drop less than 0.92\% .

\begin{table}[h!]
\vspace*{-2mm}
\centering
\setlength{\tabcolsep}{3pt}
\scalebox{0.9}{
\begin{tabular}{l|ccc|ccc|ccc|ccc|ccc}
\toprule
\multirow{1}{*}{Datasets} & \multicolumn{3}{c}{RTE} & \multicolumn{3}{c}{MRPC} & \multicolumn{3}{c}{CoLA} & \multicolumn{3}{c}{SST-2} & \multicolumn{3}{c}{QNLI}\\
\midrule
Drop Ratio&  0.00&  0.70& 0.90& 0.00&  0.70& 0.90& 0.00&  0.70& 0.90& 0.00&  0.70& 0.90& 0.00&  0.70& 0.90\\
Speed Up ($\times$)&  1.00&  2.09& 3.42& 1.00&  2.00& 3.00& 1.00&  2.29& 3.52& 1.00&  2.20& 3.35& 1.00&  2.22& 3.53\\
Accuracy ($\%$)&  65.7&  65.9& 65.3& 99.9&  89.4& 89.7& 80.7&  82.9& 81.6& 87.5&  91.6& 91.5& 88.5&  89.9& 89.4\\
\bottomrule
\end{tabular}
}
\caption{Studying the effect of drop ratio on BERT finetuning on the GLUE benchmark.}
\vspace*{-4mm}
\label{tab:BERT_results}
\end{table}

\section{Conclusion}
The computation and time requirements of pretraining and finetuning large deep learning models are ever-increasing. In this paper, we presented \textit{SwiftLearn}, a dataset sampling technique that weighs samples dynamically during training based on how much they contribute to the model's learning enabling selection of only the important samples that matter. This results in significant computation and time savings. For example, BERT finetuning on GLUE benchmark can be done with less than 0.92\% drop in average accuracy while almost 90\% of the data is dropped. \textit{SwiftLearn} opens a lot of doors for future data-efficient training. Instead of fixing the drop ratio throughout the whole training process, it can dynamically change based on the model's learning. In addition, more advanced techniques to weight the dataset samples can be used to further improve the model's training efficiency. Due to the time and space constraints, we leave those ideas for future work.   

\bibliography{natbib}
\bibliographystyle{acl_natbib}

\end{document}